\documentclass{article}

\PassOptionsToPackage{numbers, compress}{natbib}

\usepackage[preprint]{neurips_2022}

\usepackage[textsize=tiny]{todonotes}

\usepackage[utf8]{inputenc} 
\usepackage[T1]{fontenc}    
\usepackage{hyperref}       
\usepackage{url}            
\usepackage{booktabs}       
\usepackage{amsfonts}       
\usepackage{nicefrac}       
\usepackage{microtype}      
\usepackage{xcolor}         

\usepackage{booktabs}
\usepackage{multirow}
\usepackage{array,booktabs,multirow}
\usepackage{graphicx}
\usepackage{ dsfont }
\usepackage{amsmath}
\usepackage{svg}            
\DeclareGraphicsExtensions{.pdf,.png}   
\usepackage{lipsum}         

\usepackage{amsmath}
\usepackage[linesnumbered,ruled]{algorithm2e}
\usepackage{hyperref}

\newcommand*\rot{\rotatebox{90}}

\makeatletter
\newcommand{\printfnsymbol}[1]{%
  \textsuperscript{\@fnsymbol{#1}}%
}
\makeatother

\title{Employing RASCL to generalize Dispatching rules on the Job Shop Scheduling}

\author{%
  Zangir Iklassov\thanks{Equal contribution}
  \\
  \texttt{Zangir.Iklassov@mbzuai.ac.ae} \\
  \And 
  Dmitrii Medvedev\printfnsymbol{1}
  \\
  \texttt{Dmitrii.Medvedev@mbzuai.ac.ae} \\
   \And
  Ruben Solozabal\printfnsymbol{1}\\
  \texttt{Ruben.Solozabal@mbzuai.ac.ae} \\
   \And
  Martin Tak\'a\v{c} \\
   \texttt{Takac.MT@gmail.com} \\
   \And
   $\ $ \vspace{-20pt}
   \\ 
       Mohamed bin Zayed University of Artificial Intelligence (MBZUAI)
 \\
 Masdar City,
Abu Dhabi, UAE
}

\graphicspath{ {./images/} }

\begin{document}

\maketitle

\begin{abstract}

This paper introduces a Reinforcement Learning approach to better generalize heuristic dispatching rules on the Job-shop Scheduling Problem (JSP). Current models on the JSP do not focus on generalization, although, as we show in this work, this is key to learning better heuristics on the problem. A well-known technique to improve generalization is to learn on increasingly complex instances using Curriculum Learning (CL). However, as many works in the literature indicate, this technique might suffer from catastrophic forgetting when transferring the learned skills between different problem sizes. To address this issue, we introduce a novel Reinforced Adaptive Staircase Curriculum Learning (RASCL) strategy, which dynamically adjusts the difficulty level during the learning process to revisit the worst-performing instances. This work also presents a deep learning model to solve the JSP, which is equivariant w.r.t. the job definition and is size-agnostic. Conducted experiments on Taillard's and Demirkol's instances show that the presented approach significantly improves the current state-of-the-art models on the JSP. It reduces the average optimality gap from 19.35\% to 10.46\% on Taillard's instances and from 38.43\% to 18.85\% on Demirkol's instances. Our implementation is available online~\footnote{\url{https://github.com/Optimization-and-Machine-Learning-Lab/Job-Shop/tree/main_nips}}.

\end{abstract}

\section{Introduction}
\label{section:introduction}

The Job Shop Scheduling Problem (JSP) is a combinatorial problem with vast implications for scheduling optimization of real-world tasks. It is formulated as a set of jobs, each consisting of a set of operations, to be processed on a set of heterogeneous machines in the shortest possible time. Furthermore, each operation is described with two features: the specific machine each operation shall be assigned to and the operational time it takes to complete the operation. The above formulation applies to any economic task concerned with the optimal assignment of capital goods versus means of production, i.e., education, research, manufacturing, storage, transportation, sales, etc. However, JSP is an NP-hard combinatorial problem, which usually means that solving them optimally is impractical. Thus, there are two basic approaches to finding an approximate solution for this class of problems. The first approach consists of various optimization methods, including integer programming, constraint programming, or meta-heuristic algorithms. However, those are computationally expensive methods. The second approach is to employ hand-engineered heuristics that provide flexibility to react to unexpected events in the scheduling plan.

Designing such heuristic rules is a daunting task that requires specialized knowledge of the problem. For example, Priority Dispatch Rules (PDRs) are a family of scheduling heuristics that are fast and easy to implement but computationally expensive. They are also sensitive to the initialization of the problem's instances. Therefore, researchers started to gravitate toward Reinforcement Learning (RL) to discover well-behaved domain-specific heuristics automatically. Unlike PDRs, RL uses instance information to particularize a solution strategy and has already demonstrated promising results \cite{zhang2020learning, Solozabal2020constrained}, though the problem of generalization is yet to be addressed.

In this paper we provide a nostrum for generalization on the JSP using deep RL. To this end, we design a problem-specific architecture and novel training approach that stress this direction. Our main contributions are summarized below:

\begin{itemize}
    \item This work presents a \textbf{deep learning model} to solve the JSP. Particularly, we address the JSP as a Markov Decision Process (MDP), in which the model iteratively constructs a solution based on the operations that are yet to be scheduled, as well as on the state on the problem resolution. The model is equivariant w.r.t. the job information and size-agnostic, enabling us to train the model on different problem sizes. It is key for the generalization study we perform in this paper.

    \item We also present~\textbf{Reinforced Adaptive Staircase Curriculum Learning (RASCL)}, a novel curriculum strategy to improve generalization based on flexible assignment of difficulty levels during learning process. Curriculum learning (CL) used to suffer from catastrophic forgetting when transferring the learned skills between different problem sizes. To address this issue, we propose RASCL, a learning strategy that dynamically reinforces the model on worst-performing instances. Conducted experiments show this strategy presents an improvement in the difficulty selection when compared to previous CL approaches.

\end{itemize}

The aforesaid ideas deliver an improvement in performance and generalization when compared to state-of-the-art works on JSP \cite{zhang2020learning, wang2021dynamic}. Notably, we reduce the optimality gap from 19.35\% to 10.46\% on Taillard's instances and from 38.43\% to 18.85\% on Demirkol's instances.

\section{Related work}

First attempts of using RL to address scheduling problems date back to the 90s~\cite{mahadevan1997self, mahadevan1998optimizing, zhang1995reinforcement}. On specific relevance is Zhang and Dietterich’s paper~\cite{zhang1995reinforcement} on allocating resources for NASA shuttle missions. As it was reflected in the literature, one of the advantages of using RL for such a purpose is that it can be seamlessly used for static, dynamic~\cite{gabel2008adaptive, aydin2000dynamic} or stochastic variants of the problem. However, due to the complexity of the problems addressed, it could only be applied to small instances of the problem, limiting its applicability significantly.

The rise of deep learning allowed to extrapolate this technique to more realistic problem instances. Prior works particularized deep neural networks, e.g., Pointer Networks~\cite{vinyals2015pointer}, to learn in combinatorial spaces. In ~\cite{bello2016neural}, deep RL was implemented for the first time to learn \textit{end-to-end} solutions to combinatorial problems. The authors used the Pointer Network in an actor-critic architecture to address the Travelling Salesman Problem. Further studies implemented Transformer networks~\cite{deudon2018learning, vaswani2017attention, kool2018attention}. However, all these works share in common that they are based on sequence-to-sequence models, where the complete solution is output at once. Thereby, heavy sampling and searching techniques were required at interference to improve the solution. Our approach is in line with~\cite{nazari2018reinforcement}, where the Vehicle Routing Problem is described as a Markov Decision Process, and the solution is iteratively constructed based on sequential decisions.

In the particular case of the JSP, several techniques have been applied to learn on it. Imitation learning was used in~\cite{ingimundardottir2018discovering} to learn from optimal solutions on training instances that were labeled using a Mixed-Integer Programming (MIP) solver. RL has also been applied to the problem, e.g., to select pre-defined candidate PDRs according to the scheduling conditions~\cite{aydin2000dynamic, lin2019smart}. Other works have addressed the problem from a multi-agent perspective, e.g., in cooperative manufacturing ~\cite{gabel2008adaptive,waschneck2018optimization} where each agent controls a production line. Moreover, numerous examples of scheduling in many application domains, including manufacturing~\cite{lin2019smart,wang2021dynamic}, distributed computing~\cite{mao2019learning, zhang2020rlscheduler,sun2021deepweave} or supply chains. Despite the effort, many of these approaches do not beat heuristics. In addition, a major limitation in many of these works is that the state representation is hard-bounded by some factors (e.g., size of jobs or number of operations to consider), not enabling to scale the solution to arbitrary problem sizes. 

Zhang et al. ~\cite{zhang2020learning} presented a size-agnostic model that allows generalizing on different instance sizes. They formulate the JSP as a disjunctive graph and use a high discriminative Graph Isomorphism Network (GIN) to embed the states in the resolution procedure. They prove to capture raw features from small problem instances and manage to successfully extrapolate to much larger instances. Even though graph neural networks have shown unprecedented success embedding not-euclidean spaces, this embedding does not capture the most relevant information to design constructive heuristics on scheduling problems and the features of the remaining operations.

One of the key features that improve the generalization capabilities of a model is the use of CL; this aspect has been pointed out several times in the literature. E.g.~\cite{zheng2019manufacturing} employs transfer learning to reconstruct the trained policies on problems of different sizes. However, policy transfer is still relatively costly and inconvenient. Also,~\cite{liu2020actor} uses lifelong learning, where an agent will not only learn to optimize one specific problem instance but reuse what it has learned from previous instances. They also proposed a parallel training method that combines asynchronous updates with a deep deterministic policy gradient to speed up model training. In \cite{lisicki2020evaluating}, the authors use an improved CL strategy with an adaptive staircase mechanism, where, at each iteration, the model can change the difficulty level, i.e., go the previous level, stay at the same level, or advance to the next one. In our work, we build on top of this idea, tracking the model's behavior at each stage, allowing it to jump to the worst-performing levels.

\section{Method}

\label{section:method} 

\paragraph{Problem formulation.} In the deterministic JSP, a finite set of $n$ jobs $\{J_i\}_{i=1}^n$ are to be processed on a finite set of $m$ machines $\{M_k\}_{k=0}^m$. Each job $J_i$ consists of a sequence of $m$ operations $O_{i,1}\rightarrow ...\rightarrow O_{i,m}$, that have to be processed in a predetermined order. For each operation $O_{i,j}$, the machine $M_{k}$ assigned for a given processing time $D_{i,j}$, thus constituting the definition of the JSP instance. This problem aims to define the scheduling order of the operations such that the total execution time (makespan) is minimized. 

\paragraph{Constraints.} JSP model operates under several constrains listed below:
\begin{itemize}
    \item no-overlap constraint, determining that each machine shall process only one operation at a time;
    \item non-preemptive constraint, stating that once the processing of any operation is initiated, it shall not be interrupted before completion;
    \item precedence constraint, establishing the order of operations inside a job $J_i$, where operation $O_{i,j+1}$ shall not be scheduled until the previous operation $O_{i,j}$ of the job $J_i$ is completed.
\end{itemize}

\subsection{Learning the policy}

The complete schedule consists of $n \cdot m$ dispatches that must be assigned sequentially. To this end, we build an autoregressive model that, given an instance of the problem $x$, iteratively constructs the solution, and the operations are dispatched one at a time according to the scheduling policy. This is, at a decision step $t$, the model $\pi_\theta(a|s_t,x_\beta)$ takes the instance definition and the state $s_t$ as an input on the resolution process and outputs the probability distribution for taking action $a_t$ at that time. The selected operation is scheduled, and the next state $s_{t+1}$ is obtained. The process repeats until all pending operations are scheduled. At that point, the solution $y$ is defined as the sequence of selected actions $a_1,a_2,...,a_{n \cdot m}$.

With the aim of the model to minimize the makespan denoted as $T(x | \pi_\theta)$, we define the reward function at step $t$ as the temporal difference $\tau$ of two successive states:

\begin{equation}
    R_t(s_t, a_t) = \tau(s_{t+1}) - \tau(s_t) \ ,
\end{equation}

so that the makespan is comprised by the sum of collected rewards:

\begin{equation}
\label{makespan}
    T(x  | \pi_\theta ) = \sum_t R_t(s_t, a_t) \ .
\end{equation}

Using~\ref{makespan}, the model $\pi_{\theta}$ may now be expressed via the expected total processing time as follows:

\begin{equation}
\label{model_pi}
    \pi_{\theta}(a_t | s_t, x) = \arg\min_{\pi} \mathds{E} \big[ T(x | \pi_\theta) \big].
\end{equation}

Furthermore, parameter vector $\theta$ can be inferred by resorting to policy gradient methods of Reinforce Algorithm in~\cite{williams1992simple} and defining the objective function for training the model as a negative expected total reward for the policy $\pi_{\theta}$:

\begin{equation}
    \mathcal{J}^{(\pi)}(\theta) = - \mathds{E} \big[ T^{(\pi)}(x) \big] .
\end{equation}

Following the common variation of \textit{Policy Gradient Theorem}~\cite{sutton1999policy}, the gradient of the objective function can be re-written as 

\begin{equation}
    \nabla_\theta{\hat{\mathcal{J}}^{(\pi)}}(\theta) \approx - \dfrac{1}{B} \sum_{\beta=1}^{B}\sum_{t=1}^{n\cdot m} \bigg( (G(s_t,x_\beta) - b_\phi(s_t,x_\beta))  \cdot \nabla_\theta \log{\pi_\theta(a_t | s_t, x_\beta)} \bigg),
\end{equation}

where $B$ denotes the total number of batches, ${b_\phi}(s_t, x_\beta)$ is the \textit{baseline} - the estimate of the value function at state $s_t$ of the problem instance $x_\beta$, and $G(s_t,x_{\beta}) $, called \textit{return}, is the actual cumulative reward starting at time step $t$ and till all $n \cdot m$ operations are schedule:

\begin{equation}
    G(s_t,x_{\beta}) = \sum_{t}^{n \cdot m}R_t(s_t, a_t).
\end{equation}

Additionally, to reduce the variance of the gradients and, therefore, to converge faster, the actor-critic mechanism~\cite{konda1999actor, bhatnagar2007incremental} is introduced with the loss function $L$ parameterized by $\phi$ and defined as 

\begin{equation}
\label{critic_loss}
    L(\phi) = \dfrac{1}{B} \sum_{{\beta}=1}^{B} \sum_{t=1}^{n\cdot m} ||b_\phi(s_t,x_{\beta}) - G(s_t,x_{\beta}) ||^2 \ .
\end{equation}

\subsection{Inference strategies}

So far, we have defined the policy function $\pi_\theta$ providing the distribution of possible actions at each step, as well as the learning algorithm. However, as reflected in the literature~\cite{bello2016neural, kool2018attention}, many works rely on decoding strategies at inference to improve the greedy algorithm's results. In this paper, we consider the following strategies:

\paragraph{Sampling.} In sampling, strategy actions are randomly chosen from the policy distribution, i.e. $a_t \sim \pi_\theta(\cdot | s_t, x_b)$. This way, the proposed actions for the same state and instance may differ. In order for the \textit{Sampling} strategy to outperform \textit{Greedy}, it requires fine-tuning for some number of training iterations.

\paragraph{POMO.} Policy Optimization with Multiple Optima (POMO) incorporates both ideas of \textit{Greedy} and \textit{Sampling}. At the initial state $s_0$, it rolls out several actions $\{a_0^{(1)}, a_0^{(2)} , ... \}$ creating several possible trajectories $\{s_{1}^{(1)}, s_{ 1}^{(2)} , ... \}$, and then developing these trajectories \textit{greedily}. In \cite{kwon2020pomo}, authors demonstrate that, for some combinatorial problems, this approach delivers better results when compared to \textit{Sampling}.

\paragraph{Beam Search.} This strategy~\cite{joshi2019efficient, wang2021dynamic} greedily chooses $k$ actions $\{a_0^{(1)}, ..., a_0^{(k)} \}$ at the initial state $s_0$, thus forming $k$ trajectories $\{s_{1}^{(1)},..., s_{ 1}^{(k)} \}$. Then, at each trajectory, new set of $k$ actions is rolled out, and the total number of available actions becomes $k$ × $k$. To keep the number of trajectories being explored from exploding, the likelihood of each action is calculated, and the strategy greedily chooses the next $k$ actions with the highest likelihood. This way, \textit{Beam Search} explores $k$ most-likely trajectories at-each-step.

\section{Architecture}


\label{section:arch_top_view} 
\begin{figure}[t]
  \centering
  \makebox[\textwidth]{\includegraphics[width=4.5cm]{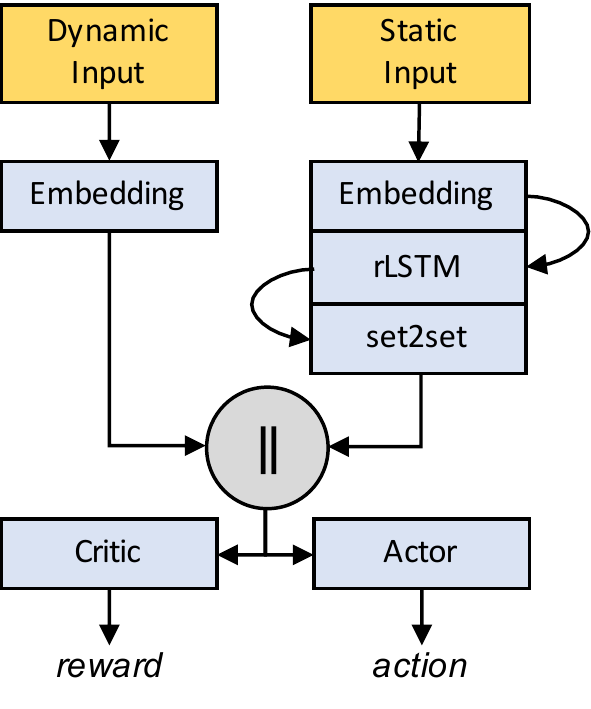}}
  \caption{Proposed RL architecture for JSP consists of deep learning recurrent preprocessing framework for complex dual input to actor-critic neural networks. Each branch of the input lands on a dedicated fully connected layer, thus allowing the model to be size-agnostic with respect to the problem's instance.}
  \label{arch_top_view}
\end{figure}

\label{section:arch_rLSTM} 
\begin{figure}[t]
  \centering
  \makebox[\textwidth]{\includegraphics[width=6cm]{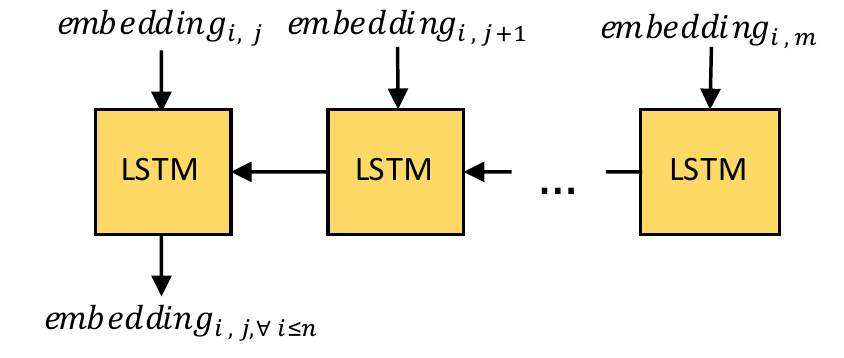}}
  \caption{Reverse LSTM receives static information in the form of multidimensional embeddings about all the operations constituting job $J_i$, and propagates this information backwards from the last operation with index $m$ to the current operation with index $j$.}
  \label{arch_rLSTM}
\end{figure}

\paragraph{Top view of architecture.} The proposed architecture is shown on Figure~\ref{arch_top_view}. It comprises of deep learning input preprocessing with the following policy inferring by actor-critic networks. The input of the model diverges into two branches - static and dynamic. Static input contains information on all operational blocks within a given job $J_i$ and is embedded in the next-to-be-scheduled operation by reversed Long Short-Term Memory network (rLSTM) \cite{hochreiter1997long}. As shown on Figure~\ref{arch_rLSTM}, rLSTM propagates the information from the ultimate future operation with index $m$ of a job $J_i$ backwards to the operation in hands with index $j$. Meanwhile, dynamic input carries information about current state of the environment: operations being currently processed, status of machines and remaining processing times. Finally, multidimensional embeddings of both branches are concatenated and passed into actor-critic networks to calculate reward of the current state and suggest the next dispatch. 

\paragraph{Job encoding and set2set.} Considering that scheduling an operation at any given time depends only on the successive operations, we employ reverse LSTM shown in Fig.~\ref{arch_rLSTM}, which propagates information starting from the last operation in this job till the current operation in hands. In order for the model to be size-agnostic, it operates on multidimensional embeddings of actual 3-dimensional operations. The output of the reverse LSTM is passed into \textit{set2set} module~\cite{vinyals2015order}, which is particularly useful to aggregate the information from all embeddings coming out from rLSTM. This module neglects the positioning indices of operations within their jobs, thus allowing to circulate updates between jobs.

\section{Improving generalization using Curriculum Learning}

\label{section:cl} 

Curriculum Learning is a class of training methodologies for policy learning using data of increasing complexity. CL  was adopted from social and educational systems to build a curricula~\cite{wang2021survey}, and offers several ways of constructing a difficulty staircase, as well as defining rules for climbing it. In this work, we consider the following most popular strategies:

\paragraph{Incremental Curriculum Learning (ICL).} Incremental CL sequentially trains the model on increasing difficulty levels. In the case of the JSP, we learn a separate model for each problem size for a fixed number of learning steps. However, the drawback of such straightforward strategy is the model's tendency to catastrophic forgetting when trained on JSP instances of consecutive grades of complexity \cite{lisicki2020evaluating}. Moreover, the number of models eventually shall be equal to the number of relevant problem sizes instead of having a single universal policy.

\paragraph{Uniform Curriculum Learning (UCL).} In uniform CL, at each iteration the model selects a problem of a random size from uniformly distributed difficulty levels. This way the model is randomly exposed to various training levels and the policy stays inherent to different problem sizes. However,  based on work of \cite{lisicki2020evaluating}, learning on small-size tasks is more accessible rather than sampling each training level from a uniform distribution of all sizes of interest, which eventually may not be the most helpful learning strategy.

\paragraph{Adaptive Staircase Curriculum Learning (ASCL).} Adaptive staircase is introduced to RL in~\cite{lisicki2020evaluating}. This strategy is based on introducing limited amount of flexibility when choosing the next JSP size for training. Specifically, the model initializes policy learning process on the lowest difficulty level available $l = l_{min}$. Then, at every $i^{th}$ iteration, the adaptive strategy, based on observed rewards, chooses one of three possible scenarios:

\begin{itemize}
    \item to advance training process to the next problem size $l \gets l + 1 $;
    \item carry on training at the same level $l$;
    \item take one step back $l \gets l - 1, \ if \ l > l_{min}$.
\end{itemize}

Such approach allows the model to start learning from the simplest policies and then sequentially increase difficulty level. Furthermore, ASCL solves problems of incremental and uniform learning strategies. However, the strategy compares model's behavior only for adjacent difficulty levels with the step difference of $1$. This approach does not estimate the model's generalization performance on other levels.

\paragraph{Reinforced Adaptive Staircase CL (RASCL)} In this work we propose reinforcement methodology for adaptive staircase learning strategy as a remedy for improving generalization of RL model on JSP. Similar to ASCL, the agent's training is initiated at the lowest difficulty level $l = l_{min}$ with the aim to make its way to the top level $l = l_{max}$. However, the limitation of ASCL is inherent in its ability to make transition only to the immediate upper or lower neighbouring states, which is often not enough to overcome the policy forgetting issue on large problem sizes.

To tackle the problem, RASCL~\ref{RASCL_Algorithm} proposes further improvement of ASCL. Specifically, every $i^{th}$ iteration future rewards are compared to the ones of optimal solution. Numerically, RASCL algorithm normalizes optimality gaps $g(l)$ for each difficulty level $l' \leq l$ and converts them into probabilities. This way, bigger gaps are assigned with higher probabilities, while each one of them corresponds to a specific problem size. The agent continues to learn at level $l$ for $u$ iterations, and afterwards the gap to optimal $g(l)$ is compared to a threshold value $t_{opt}$. In case the gap closes, RASCL choose the next problem size for training in adaptive staircase fashion $l \gets l+1$. Otherwise, every $b$ steps the reinforced adaptive algorithm samples a smaller level $l'$ from the distribution of gaps.

The proposed algorithm reinforces freshly learnt policy by revisiting and anchoring the most problematic difficulty levels. Unlike ASCL, it allows to backtrack to all the problem sizes which the agent has already visited. Being built on the basis of the former, RASCL puts the model's focus on those sizes of JSP, where it behaves the worst, at the same time keeping some flexibility and allowing the model to revisit any size.


\begin{algorithm}
    \caption{Reinforced Adaptive Curriculum Learning (RASCL) algorithm.}
    \label{RASCL_Algorithm}
    
    \SetKwInOut{Input}{Input}
    \SetKwInOut{Output}{Output}

    \Input{ dataloader $train\_data$, dataloader $test\_data$, non-negative float optimality threshold $t_{opt}$, positive integers u, b and $l_{max}$, initialized network parameters $\theta \leftarrow \theta_0$ and non-negative integer $l \leftarrow l_{0}$ }
    
    \While{$l \leq l_{max}$}{
        \For{i=0,1,...}{
            \textbf{Update g} array of percentage gaps with optimal solutions of $test\_data$ \\
            \If{i \ \% \ u == 0}{ 
              \eIf{$g[l] \leq t_{opt}$}{
                \textbf{Update $l \gets l+1$} 
              }{\If{i\%b == 0}{
                \textbf{Sample $l$} from uniform distribution of $g$ probabilities
                }
              }
             }

        \textbf{Get current size} following random distribution of $g$ \\
        \textbf{Update $\theta$} on current size data from $train\_data$ \\ 
        }
    }
    \Return{trained parameters $\theta$}
\end{algorithm}



\begin{table}[t!]
\centering
\scriptsize
\caption{Base learning on Taillard's instances. Columns represent different models trained on certain size, rows represent Taillard's data sizes. \textbf{Objective} shown as an average total time of schedules for a given size and \textbf{Gap} as an average percentage difference from optimal solutions (the less the gap the better result is)}
\scalebox{1.2}{
\begin{tabular}{lll|lllll}
\toprule
 & 
\multicolumn{1}{c}{\textbf{Instances}} &
 &
\multicolumn{1}{c}{\textbf{(15$\times$15)}} &   
\multicolumn{1}{c}{\textbf{(20$\times$15)}} &  
\multicolumn{1}{c}{\textbf{(20$\times$20)}} &  
\multicolumn{1}{c}{\textbf{(30$\times$15)}} &
\multicolumn{1}{c}{\textbf{(30$\times$20)}} \\

\midrule

 &  & Obj. & \textbf{1413.0} & 1461.8 & 1452.6 & 1470.8 & 1481.5 \\ 
 & 15$\times$15 & Gap & \textbf{(14.98\%)} & (18.97\%) & (18.23\%) & (19.7\%) & (20.59\%)\\ 
\rule{0pt}{3ex}
 &  & Obj. & \textbf{1606.6} & 1692.1 & 1692.1 & 1696.3 & 1710.2 \\ 
 & 20$\times$15 & Gap & \textbf{(17.69\%)} & (23.97\%) & (23.99\%) & (24.29\%) & (25.31\%)\\ 
\rule{0pt}{3ex}
 &  & Obj. & \textbf{1898.1} & 1964.5 & 1958.4 & 1989.8 & 2029.4 \\ 
 & 20$\times$20 & Gap & \textbf{(17.37\%)} & (21.49\%) & (21.11\%) & (23.04\%) & (25.47\%)\\ 
\rule{0pt}{3ex}
 &  & Obj. & \textbf{2153.0} & 2251.8 & 2256.2 & 2275.3 & 2276.2 \\ 
 & 30$\times$15 & Gap & \textbf{(20.41\%)} & (25.86\%) & (26.14\%) & (27.19\%) & (27.27\%)\\ 
\rule{0pt}{3ex}
 &  & Obj. & \textbf{2375.6} & 2517.7 & 2503.8 & 2516.6 & 2540.2 \\ 
 & 30$\times$20 & Gap & \textbf{(21.92\%)} & (29.22\%) & (28.52\%) & (29.19\%) & (30.4\%)\\ 
\midrule 
\rule{0pt}{3ex}
 &  & Obj. & \textbf{3207.0} & 3356.7 & 3362.6 & 3336.4 & 3346.7 \\ 
 & 50$\times$15 & Gap & \textbf{(15.67\%)} & (21.07\%) & (21.27\%) & (20.33\%) & (20.69\%)\\ 
\rule{0pt}{3ex}
 &  & Obj. & \textbf{3297.4} & 3476.0 & 3473.8 & 3496.2 & 3518.0 \\ 
 & 50$\times$20 & Gap & \textbf{(15.95\%)} & (22.25\%) & (22.15\%) & (22.96\%) & (23.73\%)\\ 
\rule{0pt}{3ex}
 &  & Obj. & \textbf{5879.9} & 6120.9 & 6110.1 & 6153.5 & 6145.5 \\ 
 & 100$\times$20 & Gap & \textbf{(9.58\%)} & (14.06\%) & (13.86\%) & (14.67\%) & (14.52\%)\\

\bottomrule
\end{tabular}
}
\vskip 5pt
\label{tab:table_base_summary}
\end{table}
\raggedbottom

\section{Experimentation}

\label{section:experiments}

\paragraph{Datasets.} We train and evaluate our model on scheduling instances of below sizes generated from Taillard’s \cite{taillard_benchmarks_1993} and Demirkol's (DMU) \cite{demirkol_benchmarks_1998} instances: 6 × 6, 10 × 10, 15 × 15, 20 × 15, 20 × 20, 30 × 20 and 30 × 15. To test a generalization of the model, we use substantially larger instances that are not included in the training set: 50 × 15, 50 × 20, and 100 × 20.

\paragraph{Baselines and base models.} For the choice of \textit{baseline models}, we resort to the analysis of priority dispatch rules (PDRs) in \cite{sels2012comparison} and choose the best performing PDRs, as well as the most popular ones in the research community: Shortest Processing Time (SPT), Minimum Ratio of Flow Due Date to Most Work Remaining (FDD/WKR), Most Work Remaining (MWKR), Most Operations Remaining (MOPNR). Among neural combinatorial solvers, we use the latest state-of-the-art results of the Graph Neural Model on public JSP benchmarks presented in \cite{zhang2020learning}. For comparison with the optimal solution, we refer to Google OR-Tools, the exact solver built on constraint programming. We also introduce the notion of \textit{base model} which has the architecture shown at Fig.~\ref{arch_top_view} and uses \textit{Sampling} strategy for selection action.

\subsection{Results} 

The implementation details can be found in Appendix~\ref{section:appendix}.

\begin{table}[!t]
\centering
\scriptsize
\caption{Incremental learning on Taillard's instances.}
\scalebox{1.2}{
\begin{tabular}{lll|lllll}
\toprule
 & 
\multicolumn{1}{c}{\textbf{Instances}} &
 &
\multicolumn{1}{c}{\textbf{(15$\times$15)}} &   
\multicolumn{1}{c}{\textbf{(20$\times$15)}} &  
\multicolumn{1}{c}{\textbf{(20$\times$20)}} &  
\multicolumn{1}{c}{\textbf{(30$\times$15)}} &
\multicolumn{1}{c}{\textbf{(30$\times$20)}} \\

\midrule

 &  & Obj. & 1413.0 & \textbf{1398.1} & 1408.5 & 1431.1 & 1435.7 \\ 
 & 15$\times$15 & Gap & (14.98\%) & \textbf{(13.76\%)} & (14.61\%) & (16.48\%) & (16.83\%)\\ 
\rule{0pt}{3ex}
 &  & Obj. & 1606.6 & \textbf{1588.1} & 1597.7 & 1653.9 & 1657.1 \\ 
 & 20$\times$15 & Gap & (17.69\%) & \textbf{(16.34\%)} & (17.06\%) & (21.17\%) & (21.41\%)\\ 
\rule{0pt}{3ex}
 &  & Obj. & 1898.1 & \textbf{1875.8} & 1892.3 & 1929.8 & 1929.4 \\ 
 & 20$\times$20 & Gap & (17.37\%) & \textbf{(15.98\%)} & (17.01\%) & (19.34\%) & (19.29\%)\\ 
\rule{0pt}{3ex}
 &  & Obj. & 2153.0 & \textbf{2123.7} & 2163.4 & 2187.7 & 2229.6 \\ 
 & 30$\times$15 & Gap & (20.41\%) & \textbf{(18.78\%)} & (20.99\%) & (22.35\%) & (24.66\%)\\ 
\rule{0pt}{3ex}
 &  & Obj. & 2375.6 & \textbf{2356.2} & 2388.4 & 2429.1 & 2455.0 \\ 
 & 30$\times$20 & Gap & (21.92\%) & \textbf{(20.94\%)} & (22.61\%) & (24.68\%) & (26.01\%)\\ 
\rule{0pt}{3ex}
 &  & Obj. & 3207.0 & \textbf{3189.0} & 3226.0 & 3264.9 & 3261.8 \\ 
 & 50$\times$15 & Gap & (15.67\%) & \textbf{(15.02\%)} & (16.35\%) & (17.77\%) & (17.64\%)\\ 
\rule{0pt}{3ex}
 &  & Obj. & 3297.4 & \textbf{3276.3} & 3326.7 & 3337.8 & 3384.4 \\ 
 & 50$\times$20 & Gap & (15.95\%) & \textbf{(15.22\%)} & (17.0\%) & (17.37\%) & (19.01\%)\\ 
\rule{0pt}{3ex}
 &  & Obj. & 5879.9 & \textbf{5851.9} & 5949.0 & 5993.4 & 6036.8 \\ 
 & 100$\times$20 & Gap & (9.58\%) & \textbf{(9.05\%)} & (10.86\%) & (11.68\%) & (12.5\%)\\

\bottomrule
\end{tabular}
}
\vskip 5pt
\label{tab:table_classic_summary}
\end{table}

\begin{figure}[!t]

    \begin{minipage}{0.48\textwidth}
        \makebox[\textwidth]{\includegraphics[width=1.05\textwidth]{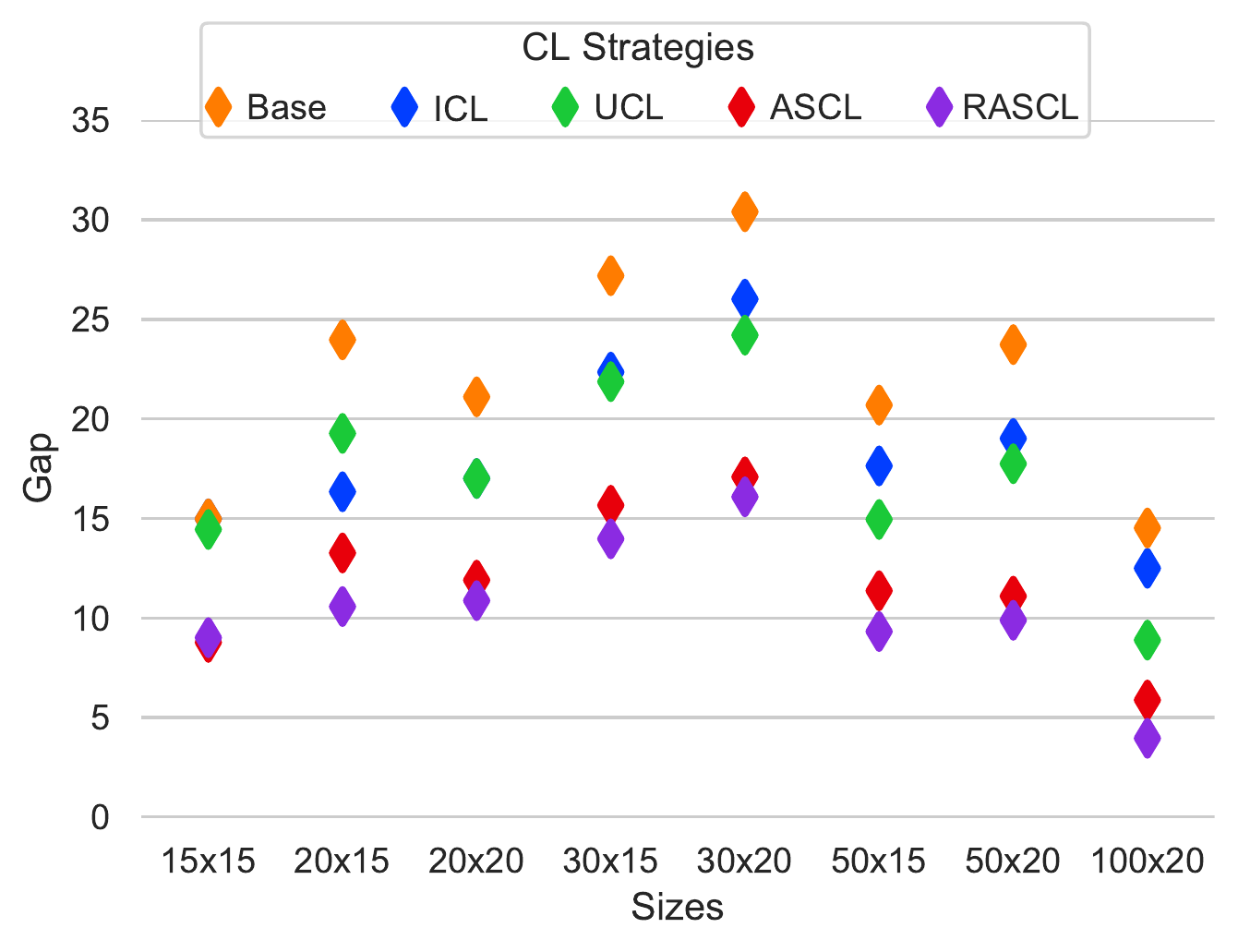}}
        \vspace{-0.50cm}
        \caption{CL and Base models results on Taillard's instances.}
        \label{fig:cl_results}
    \end{minipage}%
    \hspace{0.5cm}
    \begin{minipage}{0.48\textwidth}
        \makebox[\textwidth]{\includegraphics[width=1.05\textwidth]{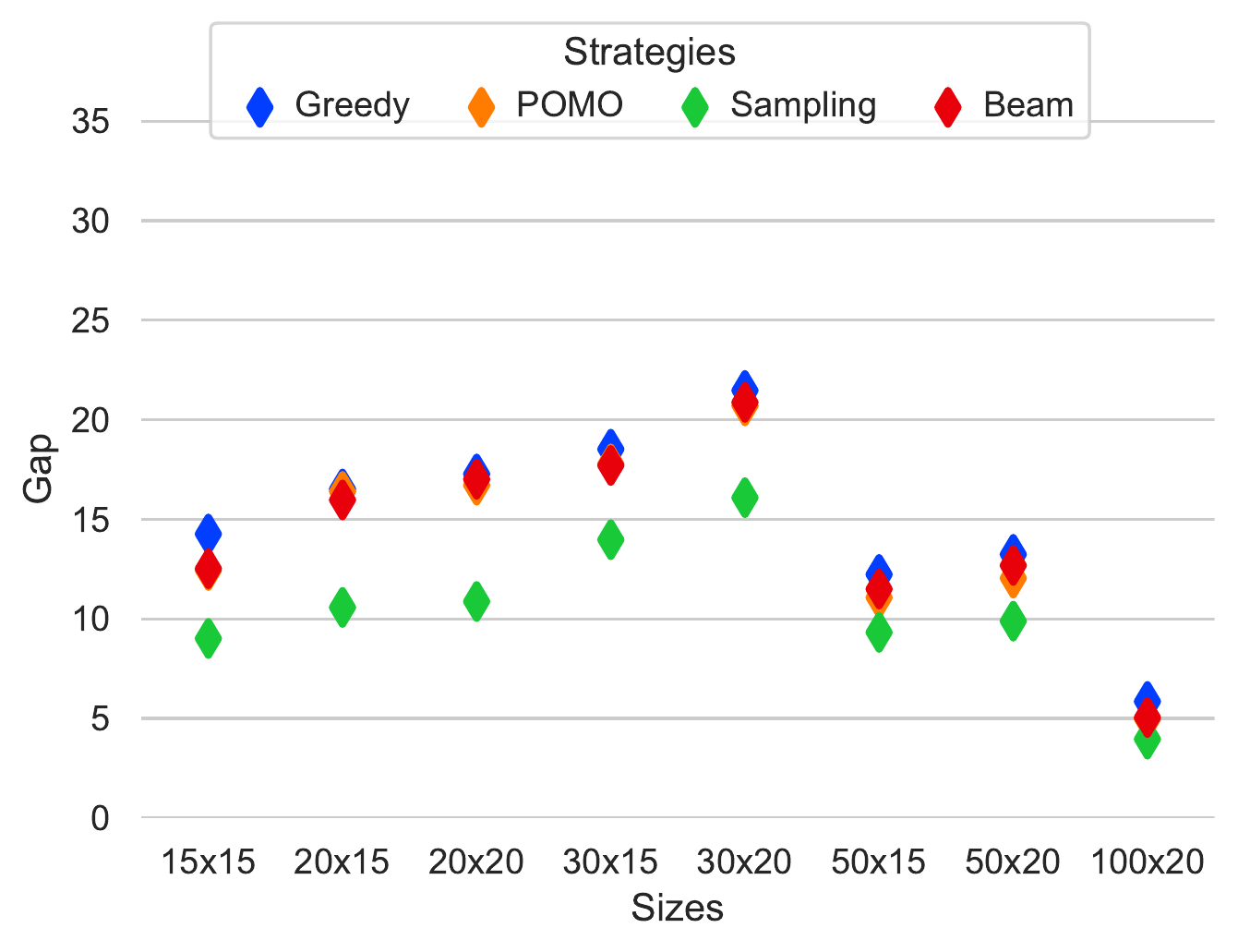}}
        \vspace{-0.50cm}
        \caption{Different sampling strategies results on Taillard's instances.}
        \label{fig:sampling_strat}
    \end{minipage} 
\end{figure}

First, we train five instances of the base model on five small-to-medium sizes: 15x15, 20x15, 20x20, 30x15, and 30x20. Each instance of the base model is dedicated to one of five problem sizes. Then, we test all base model instances on an extended set of sizes, adding 50x15, 50x20, and 100x20 to assess generalization. The results in table~\ref{tab:table_base_summary} reflect dominating performance of the base model trained on the smallest size, 15x15. It shows the smallest optimality gap on all testing sizes, from 14.98\% for 15x15 to 9.58\% for 100x20. However, the model's generalization can be improved further, as the base model trained on a small size can hardly capture all combinatorial space of action policies inherent in large-size problems.

Next, we address generalization by incorporating and comparing different CL methods. We start with ICL, for which we train five instances of the base model in the following manner: the model assigned to the 15x15 size is trained on this size only, the model assigned to 20x15 is trained sequentially on the previous size, which in this case 15x15, and on its assigned 20x15 size. The same logic applies for all sizes up to 30x20, and each learning cycle takes $n$ iterations. Table \ref{tab:table_classic_summary} shows the results of this approach. Here, the 20x15 model, being trained consecutively on 15x15 and 20x15 sizes, exhibits the best performance reducing the average gap to 9.05\% for 100x20 problem instances. ICL also improves performance for all model instances on their subsequent problem sizes compared to corresponding base models. However, the ICL approach has an obvious drawback. This drawback comes from the fact that the best performance is achieved on a smaller-size model, 20x15, meaning that the larger-size models are not learning the policies.

To address the problem of ICL, we compare it to three other types of CL (Figure~\ref{fig:cl_results}). One can see that UCL generally shows near-ICL behavior, while both adaptive and adversarial models demonstrate notably better performance. Particularly, RASCL shows the smallest optimality gap on all test sizes, except for 15x15, where ASCL slightly outperforms RASCL by 0.24\%. The adversarial approach displays the smallest optimality gap from 10.58\% for 20x15 instances to 3.96\% for 100x20 instances.

We also provide a comparison of different inference strategies on RASCL as the best-performing learning methodology. Figure \ref{fig:sampling_strat} presents a comparison of \textit{Greedy, POMO, Sampling and Beam} methods. For POMO and Beam, we use three possible trajectories. Both strategies demonstrate approximately similar results. The Greedy algorithm displays slightly worse performance because it is the special case of Beam search. The sampling strategy shows the best result on all test sizes, and, consequently, we use it for all experiments.

\begin{table}[!t]
\centering
\scriptsize
\caption{Results of Baseline and RASCL models on Taillard's insances.}
\scalebox{1.1}{
\begin{tabular}{lll|llllll}
\toprule
 & 
\multicolumn{1}{c}{\textbf{Instances}} &
 &
\multicolumn{1}{c}{\textbf{SPT}} &   
\multicolumn{1}{c}{\textbf{FDD/WKR}} &  
\multicolumn{1}{c}{\textbf{MWKR}} &  
\multicolumn{1}{c}{\textbf{MOPNR}} &
\multicolumn{1}{c}{\textbf{Zhang et al.}} &
\multicolumn{1}{c}{\textbf{RASCL}} \\

\midrule

 &  & Obj. & 1546.1 & 1808.6 & 1464.3 & 1481.3 & 1547.4 & \textbf{1339.8} \\ 
 & 15$\times$15 & Gap & (25.89\%) & (47.15\%) & (19.15\%) & (20.53\%) & (25.96\%) & \textbf{(9.02\%)}\\ 
\rule{0pt}{3ex}
 &  & Obj. & 1813.5 & 2054.0 & 1683.6 & 1686.7 & 1774.7 & \textbf{1509.3} \\ 
 & 20$\times$15 & Gap & (32.82\%) & (50.57\%) & (23.35\%) & (23.55\%) & (30.03\%) & \textbf{(10.58\%)}\\ 
\rule{0pt}{3ex}
 &  & Obj. & 2067.0 & 2387.2 & 1969.8 & 1968.3 & 2128.1 & \textbf{1793.1} \\ 
 & 20$\times$20 & Gap & (27.75\%) & (47.61\%) & (21.81\%) & (21.71\%) & (31.61\%) & \textbf{(10.87\%)}\\ 
\rule{0pt}{3ex}
 &  & Obj. & 2419.3 & 2590.8 & 2214.8 & 2195.8 & 2378.8 & \textbf{2038.1} \\ 
 & 30$\times$15 & Gap & (35.27\%) & (45.02\%) & (23.91\%) & (22.83\%) & (33.0\%) & \textbf{(13.98\%)}\\ 
\rule{0pt}{3ex}
 &  & Obj. & 2619.1 & 3045.0 & 2439.0 & 2433.6 & 2603.9 & \textbf{2261.5} \\ 
 & 30$\times$20 & Gap & (34.44\%) & (56.3\%) & (25.17\%) & (24.94\%) & (33.62\%) & \textbf{(16.09\%)}\\ 
\rule{0pt}{3ex}
 &  & Obj. & 3441.0 & 3736.3 & 3240.0 & 3254.5 & 3393.8 & \textbf{3030.8} \\ 
 & 50$\times$15 & Gap & (24.11\%) & (34.77\%) & (16.86\%) & (17.37\%) & (22.38\%) & \textbf{(9.32\%)}\\ 
\rule{0pt}{3ex}
 &  & Obj. & 3570.8 & 4022.1 & 3352.8 & 3346.9 & 3593.9 & \textbf{3125.1} \\ 
 & 50$\times$20 & Gap & (25.54\%) & (41.5\%) & (17.95\%) & (17.68\%) & (26.51\%) & \textbf{(9.89\%)}\\ 
\rule{0pt}{3ex}
 &  & Obj. & 6139.0 & 6620.7 & 5812.2 & 5856.9 & 6097.6 & \textbf{5578.9} \\ 
 & 100$\times$20 & Gap & (14.41\%) & (23.39\%) & (8.31\%) & (9.15\%) & (13.61\%) & \textbf{(3.96\%)}\\

\bottomrule
\end{tabular}
}
\label{tab:table_ta_summary}
\vskip -9.5pt
\end{table}

At this point, we incorporate RASCL for training strategy and Sampling for selection strategy to compare its overall performance versus aforesaid baselines.
Table \ref{tab:table_ta_summary} provides the comparison results on Taillard's dataset. The Reinforced Adaptive model displays robust behavior on all test sizes, improving previous best results on average by 45.94\%. This model shows a gap of 9.02\% to 13.98\% on all the sizes except for 100x20, where all compared models demonstrate relatively good performance. The reason for such observation may be the fact that Taiilard's dataset has relatively simple instances on 100x20 size.

To further test our model, we refer to DMU dataset consisting of 80 instances from 20x15 to 50x20 sizes. Table \ref{tab:table_dmu_summary} shows outperforming behavior of Reinforced Adaptive model again. On average, it improves previous best results by 50.95\%. However, absolute values of the optimality gap are higher for this data and in the range of 14.66\% to 25.42\%. It may be the result of DMU benchmark consisting of more complex instances compared to Taillard. Results of other models also confirm this fact on all test sizes.

The complete experimentation on the Taillard and DMU instances is deferred to Appendix~\ref{section:appendixB}.

\begin{table}[!t]
\centering
\scriptsize
\caption{Results of Baseline and RASCL models on DMU instances.}
\scalebox{1.1}{
\begin{tabular}{lll|llllll}
\toprule
 & 
\multicolumn{1}{c}{\textbf{Instances}} &
 &
\multicolumn{1}{c}{\textbf{SPT}} &   
\multicolumn{1}{c}{\textbf{FDD/WKR}} &  
\multicolumn{1}{c}{\textbf{MWKR}} &  
\multicolumn{1}{c}{\textbf{MOPNR}} &
\multicolumn{1}{c}{\textbf{Zhang et al.}} &
\multicolumn{1}{c}{\textbf{RASCL}} \\

\midrule

 &  & Obj. & 4951.5 & 4666.3 & 4909.9 & 4513.2 & 4215.3 & \textbf{3610.0} \\ 
 & 20$\times$15 & Gap & (64.13\%) & (53.57\%) & (62.15\%) & (49.16\%) & (38.95\%) & \textbf{(19.36\%)}\\ 
\rule{0pt}{3ex}
 &  & Obj. & 5690.5 & 5298.2 & 5489.0 & 5052.3 & 4804.5 & \textbf{4028.9} \\ 
 & 20$\times$20 & Gap & (64.57\%) & (52.52\%) & (58.16\%) & (45.17\%) & (37.74\%) & \textbf{(15.98\%)}\\ 
\rule{0pt}{3ex}
 &  & Obj. & 6306.2 & 6016.5 & 6252.9 & 5742.8 & 5557.9 & \textbf{4522.0} \\ 
 & 30$\times$15 & Gap & (62.57\%) & (54.12\%) & (60.95\%) & (47.14\%) & (41.86\%) & \textbf{(16.35\%)}\\ 
\rule{0pt}{3ex}
 &  & Obj. & 7036.0 & 6827.3 & 6925.0 & 6491.9 & 5967.4 & \textbf{5106.0} \\ 
 & 30$\times$20 & Gap & (65.91\%) & (60.09\%) & (63.16\%) & (51.97\%) & (39.48\%) & \textbf{(20.0\%)}\\ 
\rule{0pt}{3ex}
 &  & Obj. & 7601.2 & 7420.0 & 7484.2 & 7105.5 & 6663.9 & \textbf{5731.9} \\ 
 & 40$\times$15 & Gap & (55.88\%) & (51.42\%) & (52.87\%) & (44.72\%) & (35.38\%) & \textbf{(17.49\%)}\\ 
\rule{0pt}{3ex}
 &  & Obj. & 8538.1 & 8210.9 & 8460.9 & 7870.7 & 7375.8 & \textbf{6584.1} \\ 
 & 40$\times$20 & Gap & (63.0\%) & (55.52\%) & (61.11\%) & (49.22\%) & (39.38\%) & \textbf{(25.42\%)}\\ 
\rule{0pt}{3ex}
 &  & Obj. & 8975.4 & 9150.2 & 8906.0 & 8436.5 & 8179.4 & \textbf{7242.1} \\ 
 & 50$\times$15 & Gap & (50.37\%) & (52.53\%) & (48.93\%) & (40.79\%) & (36.2\%) & \textbf{(21.54\%)}\\ 
\rule{0pt}{3ex}
 &  & Obj. & 10132.8 & 9899.6 & 9807.0 & 9408.0 & 8751.6 & \textbf{7176.9} \\ 
 & 50$\times$20 & Gap & (62.2\%) & (57.26\%) & (56.4\%) & (49.61\%) & (38.86\%) & \textbf{(14.66\%)}\\ 

\bottomrule
\end{tabular}
}
\vskip 0pt
\label{tab:table_dmu_summary}
\end{table}


\section{Conclusions and Future Work}

In this work, we present a deep-RL model to automatically learn heuristic dispatching rules on the JSP. To this end, we formulate the resolution process as a Markov Decision Process, in which solutions are iteratively constructed based on intermediate states of the resolution process. In this work, we present a model that is equivariant w.r.t. the job information and size-agnostic, i.e., it enables us to train the model on instances of the JSP of different sizes. 
In order to improve the generalization of the model, this work uses Curriculum Learning. In this direction, we present a novel RASCL strategy, which dynamically adjusts the difficulty of the learning instances according to the model's performance during the learning process. Experiments on Taillard's and Demirkol's instances show that our model improves the optimality gap w.r.t. the current state-of-the-art model by 45.94\% and 50.95\%, respectively. Our results corroborate that the contributions of this paper, deep learning architecture with recurrent preprocessing and RASCL, improve the generalization on large sizes.
There are several future directions for this work. For example, we would like to remark the potential of this technology for addressing stochastic or partially observable problems, domains where traditional approaches have shown limited capabilities.

\bibliographystyle{plain}
\bibliography{refs} 

 
\appendix
\section{Appendix}
\label{section:appendix}

\textbf{Models and configurations.} In this paper, we train networks with hyperparameters being fixed for each problem size and validate each model on 1,000 randomly generated JSP instances which have the same sizes as in Taillard's dataset and remain unchanged throughout the training. For training of both Actor and Critic networks, we use Adam Optimizer with constant $10^{-4}$ learning rate and batch size of 128.

Our Base, UCL and ICL models are trained for 45000 iterations and then validated. Afterwards, the best performing model is saved. For RASCL and ASCL training, every $100_{th}$ iteration the model is tested on a test data. If during the training, the model is not increasing a difficulty level for consecutive 3000 iterations, RASCL algorithm decreases it.

Three random seeds are used during the inference and results are averaged to for final outcome. For Beam and POMO selection strategies, tree-width of 3 is used. For Sampling strategy, the sample size of 128 is used. Embedding size of $128$x$1$ is used for both static and dynamic inputs.

The repository with Python code, results, and descriptions is presented in zip archive. Detailed instructions to reproduce training, evaluation, and potting results are provided in README.md file. The model is built in Python 3.9 using PyTorch library. JSP environment is created using the gym library. Full list of all employed packages and their versions is available in Requirements.txt file stored in the repository. Training is done on NVIDIA A100 SXM 40GB GPU with 2x AMD EPYC 7742 CPUs (128 cores total with 256 threads), and 256GB RAM.

\textbf{Technical limitations} of this work relate to high demand of GPU memory when training on large size JSP instances. This potential limitation may require to decrease significantly the batch size.

\textbf{Ethical limitations.} In this work, we use synthetic data from two know publicly available datasets, Taillard's and Demirkol's scheduling instances of JSP. Furthermore, our framework is built on principals of societal responsibility and respect towards fellow researchers.

\textbf{Datasets.} One can refer to the following publicly available sources:
\begin{itemize}
    \item Taillard's dataset:   http://optimizizer.com/TA.php
    \item Demirkol's dataset:   http://optimizizer.com/DMU.php
\end{itemize}

\newpage
\section{Appendix}
\label{section:appendixB}


\begin{table}[ht!]
\centering
\scriptsize
\caption{CL strategies on Taillard's instances. Columns represent models trained using different Curriculum strategies as well as baseline Zhang's, ICL and base models, rows represent Taillard's data sizes. \textbf{Objective} shown as an average total time of schedules for a given size and \textbf{Gap} as an average percentage difference from optimal solutions (the less the gap the better result is)}
\scalebox{1.1}{
\begin{tabular}{lll|llllll}
\toprule
 & 
\multicolumn{1}{c}{\textbf{Instances}} &
 &
\multicolumn{1}{c}{\textbf{Zhang et al.}} &   
\multicolumn{1}{c}{\textbf{ICL}} &  
\multicolumn{1}{c}{\textbf{Base}} & 
\multicolumn{1}{c}{\textbf{UCL}} &  
\multicolumn{1}{c}{\textbf{ASCL}} &
\multicolumn{1}{c}{\textbf{RASCL}} \\

\midrule
\multirow{18}{*}{\rot{\textsc{Taillard}}}

 &  & Obj. & 1547.4 & 1413.0 & 1413.0 & 1406.1 & \textbf{1336.6} & 1339.8 \\ 
 & 15$\times$15 & Gap & (25.96\%) & (14.98\%) & (14.98\%) & (14.45\%) & \textbf{(8.78\%)} & (9.02\%)\\ 
\rule{0pt}{3ex}
 &  & Obj. & 1774.7 & 1588.1 & 1692.1 & 1627.7 & 1546.2 & \textbf{1509.3} \\ 
 & 20$\times$15 & Gap & (30.03\%) & (16.34\%) & (23.97\%) & (19.27\%) & (13.27\%) & \textbf{(10.58\%)}\\ 
\rule{0pt}{3ex}
 &  & Obj. & 2128.1 & 1892.3 & 1958.4 & 1892.1 & 1809.8 & \textbf{1793.1} \\ 
 & 20$\times$20 & Gap & (31.61\%) & (17.01\%) & (21.11\%) & (17.01\%) & (11.91\%) & \textbf{(10.87\%)}\\ 
\rule{0pt}{3ex}
 &  & Obj. & 2378.8 & 2187.7 & 2275.3 & 2179.3 & 2068.1 & \textbf{2038.1} \\ 
 & 30$\times$15 & Gap & (33.0\%) & (22.35\%) & (27.19\%) & (21.87\%) & (15.66\%) & \textbf{(13.98\%)}\\ 
\rule{0pt}{3ex}
 &  & Obj. & 2603.9 & 2455.0 & 2540.2 & 2419.5 & 2281.0 & \textbf{2261.5} \\ 
 & 30$\times$20 & Gap & (33.62\%) & (26.01\%) & (30.4\%) & (24.2\%) & (17.09\%) & \textbf{(16.09\%)}\\ 
\rule{0pt}{3ex}
 &  & Obj. & 3393.8 & 3261.8 & 3346.7 & 3187.5 & 3088.1 & \textbf{3030.8} \\ 
 & 50$\times$15 & Gap & (22.38\%) & (17.64\%) & (20.69\%) & (14.95\%) & (11.37\%) & \textbf{(9.32\%)}\\ 
\rule{0pt}{3ex}
 &  & Obj. & 3593.9 & 3384.4 & 3518.0 & 3348.1 & 3159.4 & \textbf{3125.1} \\ 
 & 50$\times$20 & Gap & (26.51\%) & (19.01\%) & (23.73\%) & (17.75\%) & (11.1\%) & \textbf{(9.89\%)}\\ 
\rule{0pt}{3ex}
 &  & Obj. & 6097.6 & 6036.8 & 6145.5 & 5843.1 & 5682.1 & \textbf{5578.9} \\ 
 & 100$\times$20 & Gap & (13.61\%) & (12.5\%) & (14.52\%) & (8.89\%) & (5.89\%) & \textbf{(3.96\%)}\\ 

\bottomrule
\end{tabular}
}
\vskip 5pt
\label{tab:table_cl_summary}
\end{table}

\begin{table}[!ht]
\centering
\scriptsize
\caption{Selection strategies on Taillard's instances.}
\scalebox{1.1}{
\begin{tabular}{lll|llllll}
\toprule
 & 
\multicolumn{1}{c}{\textbf{Instances}} &
 &
\multicolumn{1}{c}{\textbf{Greedy}} &   
\multicolumn{1}{c}{\textbf{POMO}} &  
\multicolumn{1}{c}{\textbf{Sampling}} &  
\multicolumn{1}{c}{\textbf{Beam}} &  \\

\midrule
\multirow{18}{*}{\rot{\textsc{Taillard}}}

 &  & Obj. & 1404.2 & 1381.9 & \textbf{1339.8} & 1382.9 \\ 
 & 15$\times$15 & Gap & (14.26\%) & (12.43\%) & \textbf{(9.02\%)} & (12.52\%)\\ 
\rule{0pt}{3ex}
 &  & Obj. & 1590.3 & 1588.6 & \textbf{1509.3} & 1582.9 \\ 
 & 20$\times$15 & Gap & (16.52\%) & (16.4\%) & \textbf{(10.58\%)} & (15.97\%)\\ 
\rule{0pt}{3ex}
 &  & Obj. & 1896.5 & 1887.5 & \textbf{1793.1} & 1892.1 \\ 
 & 20$\times$20 & Gap & (17.27\%) & (16.71\%) & \textbf{(10.87\%)} & (17.01\%)\\ 
\rule{0pt}{3ex}
 &  & Obj. & 2118.9 & 2105.0 & \textbf{2038.1} & 2104.8 \\ 
 & 30$\times$15 & Gap & (18.52\%) & (17.78\%) & \textbf{(13.98\%)} & (17.7\%)\\ 
\rule{0pt}{3ex}
 &  & Obj. & 2365.7 & 2350.8 & \textbf{2261.5} & 2354.0 \\ 
 & 30$\times$20 & Gap & (21.47\%) & (20.7\%) & \textbf{(16.09\%)} & (20.87\%)\\ 
\rule{0pt}{3ex}
 &  & Obj. & 3111.6 & 3079.1 & \textbf{3030.8} & 3090.9 \\ 
 & 50$\times$15 & Gap & (12.23\%) & (11.08\%) & \textbf{(9.32\%)} & (11.5\%)\\ 
\rule{0pt}{3ex}
 &  & Obj. & 3219.6 & 3185.9 & \textbf{3125.1} & 3203.4 \\ 
 & 50$\times$20 & Gap & (13.24\%) & (12.05\%) & \textbf{(9.89\%)} & (12.68\%)\\ 
\rule{0pt}{3ex}
 &  & Obj. & 5680.9 & 5633.3 & \textbf{5578.9} & 5637.5 \\ 
 & 100$\times$20 & Gap & (5.86\%) & (4.97\%) & \textbf{(3.96\%)} & (5.05\%)\\ 

\bottomrule
\end{tabular}
}
\vskip 5pt

\label{tab:table_selection_summary}
\end{table}

\begin{table}[!ht]
\centering
\scriptsize
\caption{Selection strategies on Taillard's instances.}
\scalebox{0.9}{
\begin{tabular}{lllllll}
\toprule
{}&
\multicolumn{1}{c}{\textbf{Instance}} & 
\multicolumn{1}{c}{\textbf{Greedy}} &   
\multicolumn{1}{c}{\textbf{POMO}} &  
\multicolumn{1}{c}{\textbf{Sampling}} &  
\multicolumn{1}{c}{\textbf{Beam}} & 
\multicolumn{1}{c}{\textbf{UB}} \\

\midrule
\multirow{10}{*}{\rot{15$\times$15}}
 & TA1 & 1445 (17.38\%) & 1445 (17.38\%) & \textbf{1379} \textbf{(12.02\%)} & 1421 (15.43\%) & 1231* \\ 
 & TA2 & 1405 (12.94\%) & 1405 (12.94\%) & \textbf{1325} \textbf{(6.51\%)} & 1390 (11.74\%) & 1244* \\ 
 & TA3 & 1469 (20.61\%) & 1344 (10.34\%) & \textbf{1338} \textbf{(9.85\%)} & 1405 (15.35\%) & 1218* \\ 
 & TA4 & 1304 (10.98\%) & 1304 (10.98\%) & \textbf{1275} \textbf{(8.51\%)} & 1304 (10.98\%) & 1175* \\ 
 & TA5 & 1345 (9.89\%) & 1345 (9.89\%) & \textbf{1303} \textbf{(6.45\%)} & 1326 (8.33\%) & 1224* \\ 
 & TA6 & 1360 (9.85\%) & 1329 (7.35\%) & \textbf{1325} \textbf{(7.03\%)} & 1326 (7.11\%) & 1238* \\ 
 & TA7 & 1426 (16.22\%) & 1426 (16.22\%) & \textbf{1332} \textbf{(8.56\%)} & 1400 (14.1\%) & 1227* \\
\bottomrule
\end{tabular}
}
\label{tab:table_selection_strats}
\end{table}

\begin{table}[!ht]
\centering
\scriptsize
\caption{ICL on Taillard's instances.}
\scalebox{0.9}{
\begin{tabular}{llllllll}
\toprule
{}&
\multicolumn{1}{c}{\textbf{Instance}} & 
\multicolumn{1}{c}{\textbf{(15$\times$15)}} &   
\multicolumn{1}{c}{\textbf{(20$\times$15)}} &  
\multicolumn{1}{c}{\textbf{(20$\times$20)}} &  
\multicolumn{1}{c}{\textbf{(30$\times$15)}} &
\multicolumn{1}{c}{\textbf{(30$\times$20)}} &
\multicolumn{1}{c}{\textbf{UB}} \\

\midrule
\multirow{10}{*}{\rot{15$\times$15}}
 & TA1 & 1475 (19.82\%) & 1457 (18.36\%) & 1458 (18.44\%) & \textbf{1453} \textbf{(18.03\%)} & 1477 (19.98\%) & 1231* \\ 
 & TA2 & \textbf{1401} \textbf{(12.62\%)} & 1414 (13.67\%) & 1410 (13.34\%) & 1424 (14.47\%) & 1426 (14.63\%) & 1244* \\ 
 & TA3 & \textbf{1393} \textbf{(14.37\%)} & 1397 (14.7\%) & 1413 (16.01\%) & 1467 (20.44\%) & 1447 (18.8\%) & 1218* \\ 
 & TA4 & 1340 (14.04\%) & \textbf{1338} \textbf{(13.87\%)} & 1351 (14.98\%) & 1412 (20.17\%) & 1374 (16.94\%) & 1175* \\ 
 & TA5 & 1390 (13.56\%) & \textbf{1354} \textbf{(10.62\%)} & 1404 (14.71\%) & 1433 (17.08\%) & 1388 (13.4\%) & 1224* \\ 
 & TA6 & 1388 (12.12\%) & \textbf{1382} \textbf{(11.63\%)} & 1389 (12.2\%) & 1401 (13.17\%) & 1414 (14.22\%) & 1238* \\ 
 & TA7 & 1425 (16.14\%) & \textbf{1370} \textbf{(11.65\%)} & 1374 (11.98\%) & 1408 (14.75\%) & 1479 (20.54\%) & 1227* \\
\bottomrule
\end{tabular}
}
\label{tab:table_classic_cl}
\end{table}

\begin{table}[!ht]
\centering
\scriptsize
\caption{Base learning on Taillard's instances.}
\scalebox{0.9}{
\begin{tabular}{llllllll}
\toprule
{}&
\multicolumn{1}{c}{\textbf{Instance}} & 
\multicolumn{1}{c}{\textbf{(15$\times$15)}} &   
\multicolumn{1}{c}{\textbf{(20$\times$15)}} &  
\multicolumn{1}{c}{\textbf{(20$\times$20)}} &  
\multicolumn{1}{c}{\textbf{(30$\times$15)}} &
\multicolumn{1}{c}{\textbf{(30$\times$20)}} &
\multicolumn{1}{c}{\textbf{UB}} \\

\midrule
\multirow{10}{*}{\rot{15$\times$15}}
 & TA1 & 1475 (19.82\%) & 1492 (21.2\%) & 1466 (19.09\%) & \textbf{1443} \textbf{(17.22\%)} & 1477 (19.98\%) & 1231* \\ 
 & TA2 & \textbf{1401} \textbf{(12.62\%)} & 1429 (14.87\%) & 1437 (15.51\%) & 1463 (17.6\%) & 1445 (16.16\%) & 1244* \\ 
 & TA3 & \textbf{1393} \textbf{(14.37\%)} & 1476 (21.18\%) & 1470 (20.69\%) & 1484 (21.84\%) & 1481 (21.59\%) & 1218* \\ 
 & TA4 & \textbf{1340} \textbf{(14.04\%)} & 1437 (22.3\%) & 1435 (22.13\%) & 1470 (25.11\%) & 1513 (28.77\%) & 1175* \\ 
 & TA5 & \textbf{1390} \textbf{(13.56\%)} & 1409 (15.11\%) & 1451 (18.55\%) & 1464 (19.61\%) & 1486 (21.41\%) & 1224* \\ 
 & TA6 & \textbf{1388} \textbf{(12.12\%)} & 1408 (13.73\%) & 1413 (14.14\%) & 1444 (16.64\%) & 1421 (14.78\%) & 1238* \\ 
 & TA7 & 1425 (16.14\%) & 1453 (18.42\%) & \textbf{1398} \textbf{(13.94\%)} & 1463 (19.23\%) & 1482 (20.78\%) & 1227* \\
\bottomrule
\end{tabular}
}
\label{tab:table_base}
\end{table}

\begin{table}[!ht]
\centering
\scriptsize
\caption{CL strategies on Taillard's instances.}
\scalebox{0.9}{
\begin{tabular}{lllllllll}
\toprule
{}&
\multicolumn{1}{c}{\textbf{Instance}} & 
\multicolumn{1}{c}{\textbf{Zhang et al.}} &   
\multicolumn{1}{c}{\textbf{ICL}} &  
\multicolumn{1}{c}{\textbf{Base}} & 
\multicolumn{1}{c}{\textbf{UCL}} &  
\multicolumn{1}{c}{\textbf{ASCL}} &
\multicolumn{1}{c}{\textbf{RASCL}} &
\multicolumn{1}{c}{\textbf{UB}} \\

\midrule
\multirow{10}{*}{\rot{15$\times$15}}
 & TA1 & 1443 (17.22\%) & 1475 (19.82\%) & 1475 (19.82\%) & 1393 (13.16\%) & 1387 (12.67\%) & \textbf{1379} \textbf{(12.02\%)} & 1231* \\ 
 & TA2 & 1544 (24.12\%) & 1401 (12.62\%) & 1401 (12.62\%) & 1391 (11.82\%) & 1336 (7.4\%) & \textbf{1325} \textbf{(6.51\%)} & 1244* \\ 
 & TA3 & 1440 (18.23\%) & 1393 (14.37\%) & 1393 (14.37\%) & 1408 (15.6\%) & 1341 (10.1\%) & \textbf{1338} \textbf{(9.85\%)} & 1218* \\ 
 & TA4 & 1637 (39.32\%) & 1340 (14.04\%) & 1340 (14.04\%) & 1404 (19.49\%) & 1305 (11.06\%) & \textbf{1275} \textbf{(8.51\%)} & 1175* \\ 
 & TA5 & 1619 (32.27\%) & 1390 (13.56\%) & 1390 (13.56\%) & 1407 (14.95\%) & 1329 (8.58\%) & \textbf{1303} \textbf{(6.45\%)} & 1224* \\ 
 & TA6 & 1601 (29.32\%) & 1388 (12.12\%) & 1388 (12.12\%) & 1375 (11.07\%) & \textbf{1309} \textbf{(5.74\%)} & 1325 (7.03\%) & 1238* \\ 
 & TA7 & 1568 (27.79\%) & 1425 (16.14\%) & 1425 (16.14\%) & 1398 (13.94\%) & \textbf{1321} \textbf{(7.66\%)} & 1332 (8.56\%) & 1227* \\
\bottomrule
\end{tabular}
}
\label{tab:table_cl_comparison}
\end{table}

\begin{table}[!ht]
\centering
\scriptsize
\caption{Results on Taillard’s benchmark.}
\scalebox{0.9}{
\begin{tabular}{lllllllll}
\toprule
{}&
\multicolumn{1}{c}{\textbf{Instance}} & 
\multicolumn{1}{c}{\textbf{SPT}} &   
\multicolumn{1}{c}{\textbf{FDD/WKR}} &  
\multicolumn{1}{c}{\textbf{MWKR}} &  
\multicolumn{1}{c}{\textbf{MOPNR}} &
\multicolumn{1}{c}{\textbf{Zhang et al.}} &
\multicolumn{1}{c}{\textbf{RASCL}} &
\multicolumn{1}{c}{\textbf{UB}} \\

\midrule
\multirow{10}{*}{\rot{15$\times$15}}
 & TA1 & 1462 (18.77\%) & 1841 (49.55\%) & 1491 (21.12\%) & 1438 (16.82\%) & 1443 (17.22\%) & \textbf{1379} \textbf{(12.02\%)} & 1231* \\ 
 & TA2 & 1446 (16.24\%) & 1895 (52.33\%) & 1440 (15.76\%) & 1452 (16.72\%) & 1544 (24.12\%) & \textbf{1325} \textbf{(6.51\%)} & 1244* \\ 
 & TA3 & 1495 (22.74\%) & 1914 (57.14\%) & 1426 (17.08\%) & 1418 (16.42\%) & 1440 (18.23\%) & \textbf{1338} \textbf{(9.85\%)} & 1218* \\ 
 & TA4 & 1708 (45.36\%) & 1653 (40.68\%) & 1387 (18.04\%) & 1457 (24.0\%) & 1637 (39.32\%) & \textbf{1275} \textbf{(8.51\%)} & 1175* \\ 
 & TA5 & 1618 (32.19\%) & 1787 (46.0\%) & 1494 (22.06\%) & 1448 (18.3\%) & 1619 (32.27\%) & \textbf{1303} \textbf{(6.45\%)} & 1224* \\ 
 & TA6 & 1522 (22.94\%) & 1748 (41.2\%) & 1369 (10.58\%) & 1486 (20.03\%) & 1601 (29.32\%) & \textbf{1325} \textbf{(7.03\%)} & 1238* \\ 
 & TA7 & 1434 (16.87\%) & 1660 (35.29\%) & 1470 (19.8\%) & 1456 (18.66\%) & 1568 (27.79\%) & \textbf{1332} \textbf{(8.56\%)} & 1227* \\
\bottomrule
\end{tabular}
}
\label{tab:table_ta}
\end{table}

\begin{table}[!ht]
\centering
\scriptsize
\caption{Results on DMU benchmark.}
\scalebox{0.9}{
\begin{tabular}{lllllllll}
\toprule
{}&
\multicolumn{1}{c}{\textbf{Instance}} & 
\multicolumn{1}{c}{\textbf{SPT}} &   
\multicolumn{1}{c}{\textbf{FDD/WKR}} &  
\multicolumn{1}{c}{\textbf{MWKR}} &  
\multicolumn{1}{c}{\textbf{MOPNR}} &
\multicolumn{1}{c}{\textbf{Zhang et al.}} &
\multicolumn{1}{c}{\textbf{RASCL}} &
\multicolumn{1}{c}{\textbf{UB}} \\

\midrule
\multirow{10}{*}{\rot{20$\times$15}}
 & DMU1 & 4516 (76.2\%) & 3535 (37.92\%) & 3988 (55.6\%) & 3882 (51.46\%) & 3323 (29.65\%) & \textbf{3107} \textbf{(21.23\%)} & 2563* \\ 
 & DMU2 & 4593 (69.73\%) & 3847 (42.17\%) & 4555 (68.33\%) & 3884 (43.53\%) & 3630 (34.15\%) & \textbf{3272} \textbf{(20.92\%)} & 2706* \\ 
 & DMU3 & 4438 (62.5\%) & 4063 (48.77\%) & 4117 (50.75\%) & 3979 (45.7\%) & 3660 (34.02\%) & \textbf{3270} \textbf{(19.74\%)} & 2731* \\ 
 & DMU4 & 4533 (69.84\%) & 4160 (55.86\%) & 3995 (49.68\%) & 4079 (52.83\%) & 3816 (42.97\%) & \textbf{3138} \textbf{(17.57\%)} & 2669* \\ 
 & DMU5 & 4420 (60.79\%) & 4238 (54.17\%) & 4977 (81.05\%) & 4116 (49.73\%) & 3897 (41.76\%) & \textbf{3290} \textbf{(19.68\%)} & 2749* \\ 
 & DMU41 & 5283 (62.65\%) & 5187 (59.7\%) & 5377 (65.55\%) & 5070 (56.1\%) & 4316 (32.88\%) & \textbf{3807} \textbf{(17.21\%)} & 3248* \\ 
 & DMU42 & 5354 (57.94\%) & 5583 (64.69\%) & 6076 (79.23\%) & 4976 (46.78\%) & 4858 (43.3\%) & \textbf{4027} \textbf{(18.79\%)} & 3390* \\
\bottomrule
\end{tabular}
}
\label{tab:table_dmu}
\end{table}

\end{document}